\documentclass[conference]{IEEEtran}
\usepackage{cite}

\usepackage{amsmath,amssymb,amsfonts}
\usepackage{algorithmic}
\usepackage{graphicx}
\usepackage{textcomp}
\usepackage{xcolor}

\usepackage[most]{tcolorbox}
\definecolor{block-gray}{gray}{0.85}
\newtcolorbox{prompt}{colback=block-gray,grow to right by=0mm,grow to left by=0mm,
boxrule=0pt,boxsep=0pt,breakable}

\newtcolorbox{promptboxno}[2][]{
  colback=gray!10!white,
  colframe=gray!50!black,
  boxrule=0.5pt,
  arc=3pt,
  left=6pt,
  right=6pt,
  top=6pt,
  bottom=6pt,
  fonttitle=\bfseries,
   title={#2},
   label={#1},
}

\newcounter{promptcounter}
\newtcolorbox[auto counter, number within=section]{promptbox}[2][]{%
  colback=gray!10!white,
  colframe=gray!50!black,
  boxrule=0.5pt,
  arc=3pt,
  fonttitle=\bfseries,
  title={Prompt~\thetcbcounter: #2},
  label={#1},
}

\usepackage{xcolor}

\colorlet{punct}{red!60!black}
\definecolor{background}{HTML}{EEEEEE}
\definecolor{delim}{RGB}{20,105,176}
\colorlet{numb}{magenta!60!black}

\lstdefinelanguage{json}{
    showstringspaces=false,
    breaklines=true,
    frame=lines,
    backgroundcolor=\color{background},
    literate=
     *{0}{{{\color{numb}0}}}{1}
      {1}{{{\color{numb}1}}}{1}
      {2}{{{\color{numb}2}}}{1}
      {3}{{{\color{numb}3}}}{1}
      {4}{{{\color{numb}4}}}{1}
      {5}{{{\color{numb}5}}}{1}
      {6}{{{\color{numb}6}}}{1}
      {7}{{{\color{numb}7}}}{1}
      {8}{{{\color{numb}8}}}{1}
      {9}{{{\color{numb}9}}}{1}
      {:}{{{\color{punct}{:}}}}{1}
      {,}{{{\color{punct}{,}}}}{1}
      {\{}{{{\color{delim}{\{}}}}{1}
      {\}}{{{\color{delim}{\}}}}}{1}
      {[}{{{\color{delim}{[}}}}{1}
      {]}{{{\color{delim}{]}}}}{1},
}
\usepackage{mdframed}
\usepackage{url}
\usepackage{tabto}
\usepackage{hhline}
\usepackage{times}
\usepackage{latexsym}
\usepackage{fancyvrb}
\usepackage{booktabs}
\usepackage{tabularx}
\usepackage{multirow}
\usepackage{multicol}
\usepackage{graphicx}
\usepackage{subcaption}
\usepackage{tabularray}
\usepackage{xcolor}
\usepackage{amsmath, amssymb}
\usepackage{mdframed}
\usepackage{listings}
\usepackage{wrapfig}

\def\BibTeX{{\rm B\kern-.05em{\sc i\kern-.025em b}\kern-.08em
    T\kern-.1667em\lower.7ex\hbox{E}\kern-.125emX}}
\begin{document}

\title{
Semantically-Aware LLM Agent to Enhance Privacy in Conversational AI Services
}

\author{
\IEEEauthorblockN{Jayden Serenari}
\IEEEauthorblockA{\textit{Department of Computer Science} \\
\textit{University of Pittsburgh}\\
Pittsburgh, Pennsylvania, USA}

\and
\IEEEauthorblockN{Stephen Lee}
\IEEEauthorblockA{\textit{Department of Computer Science} \\
\textit{University of Pittsburgh}\\
Pittsburgh, Pennsylvania, USA}

}

\maketitle

\begin{abstract}
With the increasing use of conversational AI systems, there is growing concern over privacy leaks, especially when users share sensitive personal data in interactions with Large Language Models (LLMs). Conversations shared with these models may contain Personally Identifiable Information (PII), which, if exposed, could lead to security breaches or identity theft. 
To address this challenge, we present the Local Optimizations for Pseudonymization with Semantic Integrity Directed Entity Detection (LOPSIDED) framework, a semantically-aware privacy agent designed to safeguard sensitive PII data when using remote LLMs. Unlike prior work that often degrade response quality, our approach dynamically replaces sensitive PII entities in user prompts with semantically consistent pseudonyms, preserving the contextual integrity of conversations. Once the model generates its response, the pseudonyms are automatically depseudonymized, ensuring the user receives an accurate, privacy-preserving output. 
We evaluate our approach using real-world conversations sourced from ShareGPT, which we further augment and annotate to assess whether named entities are contextually relevant to the model's response. 
Our results show that LOPSIDED reduces semantic utility errors by a factor of 5 compared to baseline techniques, all while enhancing privacy.


\end{abstract}

\begin{IEEEkeywords}
privacy, LLM, Personally Identifiable Information (PII)
\end{IEEEkeywords}

\section{Introduction}
%
Conversational AI systems are rapidly being adopted across domains ranging from customer service and healthcare to creative writing and programming assistance. These systems rely on Large Language Models (LLMs) hosted on remote servers, requiring user inputs that may include potentially sensitive data, to be transmitted for inference. 
When users unknowingly share Personally Identifiable Information (PII), such as names and addresses, they expose themselves to privacy risks, regulatory liabilities, and possible data breaches~\cite{10.1145/1179601.1179608}.




\begin{figure}
    \centering
    \includegraphics[width=1\linewidth]{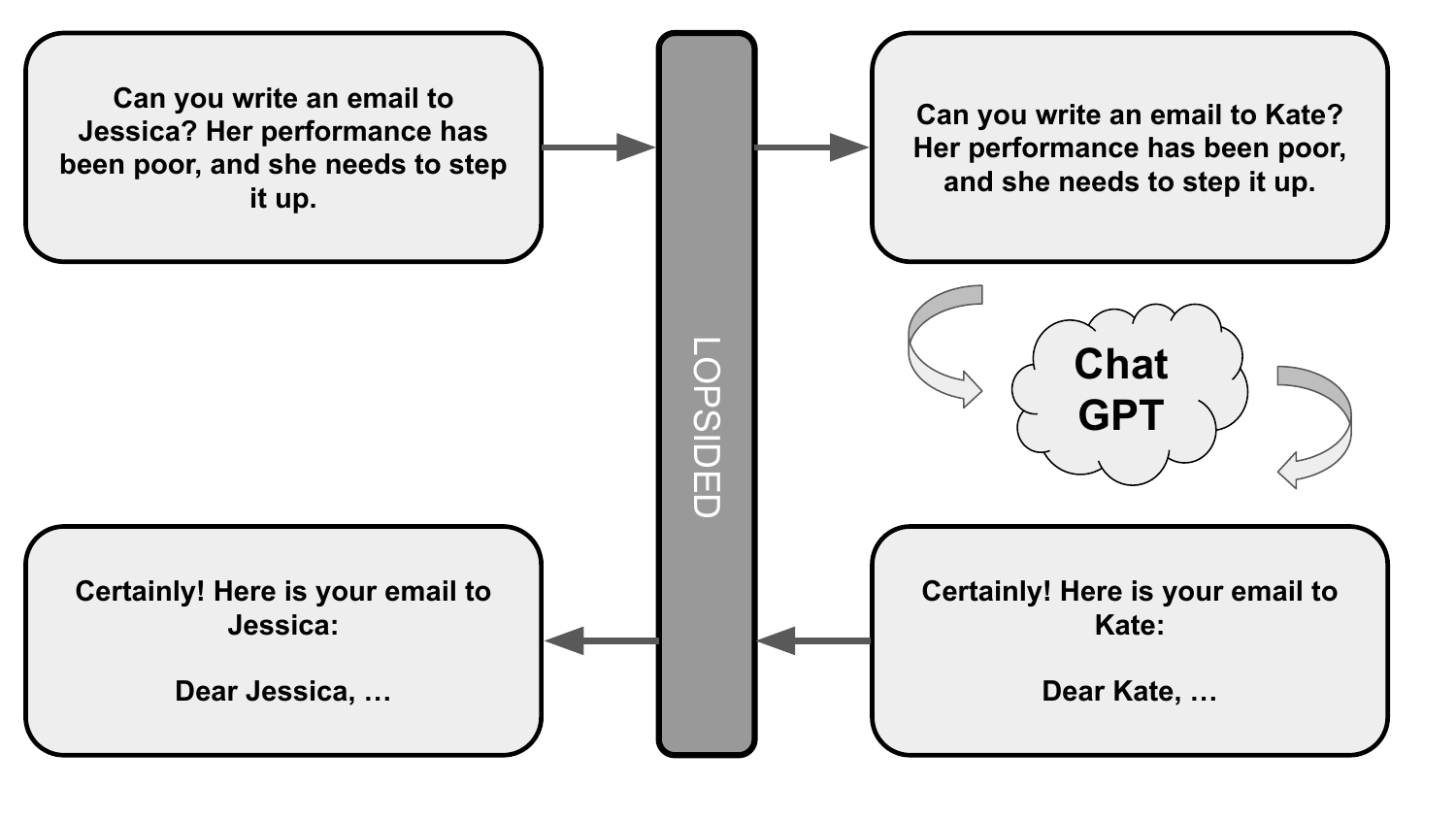}
    \caption{The LOPSIDED privacy agent system design.}
    \label{fig:design}
\end{figure}

Pseudonymization is a common approach~\cite{STUBBS2015S11} used to protect users' privacy by replacing PII with entities of the same class. For example, a city name like \textit{Chicago} might be substituted with \textit{Los Angeles} to obscure the original data while maintaining the overall structure of the input. However, if a system relies on specific details for accuracy, altering key information may lead to misleading or incorrect results. For instance, if a user asks, "What is the population of Chicago?" and the system modifies it to "What is the population of Los Angeles?", the semantic integrity of the query is compromised. This highlights a key challenge in privacy-preserving techniques --- ensuring that user data remains protected without distorting the intended \textit{semantic meaning} of their input.


More recently, LLMs have been explored for PII removal in conversational AI  services~\cite{dou2023reducing}. For example, Hide and Seek (HAS)~\cite{chen2023hideseekhaslightweight} anonymizes any PII within a prompt before it is transmitted to a cloud-based language model and then de-anonymizes the LLM's response. This approach is subject to the same pitfall as stated previously. This gap motivates a key research question: \textit{How can we preserve user privacy through pseudonymization while ensuring that LLM responses remain semantically faithful to the original query?}



Currently, all techniques rely on an \textit{all-or-nothing approach}, where all private entities are removed or none are. This can render the system unusable for users, as they may not be able to effectively use the tool if essential information is removed. Our key insight is to develop a more nuanced approach that selectively pseudonymizes PII data and generates a semantically similar response. For example, if a user asks about the weather in Palo Alto, it could be replaced with San Jose, maintaining privacy while preserving the accuracy of the response.


To address this challenge, we propose LOPSIDED, a lightweight framework that balances PII removal and semantic response preservation. Our work focuses on maximizing user privacy by locally pseudonymizing sensitive information before it is transmitted to remote LLMs. 
The privacy agent, shown in Figure~\ref{fig:design},  operates as an intermediary between the user and the cloud. It intercepts the request, pseudonymizes the prompt by replacing PII, and de-anonymizes it before presenting it to the user. This ensures that privacy-sensitive data is not exposed, while maintaining the relevance of the system's response. We note that there are situations where a replacement could completely alter the meaning. In such cases, we prioritize maintaining the utility of the response. Since pseudonymization must occur locally, we explore the use of small models that can be deployed on the user’s device. Our key contributions are as follows:

    \noindent
    \textit{LOPSIDED Design}: 
    We formulate the problem of semantic-aware privacy for conversational AI, where the goal is to pseudonymize named entities while preserving the utility of the LLM response. To address this problem, we introduce LOPSIDED, a framework which ensures that named entities can be modified without disrupting the semantic integrity of the response, making it both privacy-preserving and semantically accurate. 
    
    \noindent
    \textit{Semantic-aware Privacy Dataset}: We augment the ShareGPT dataset, which contains real-world ChatGPT conversation histories, by annotating named entities for relevance to the prompt. This results a novel 866-sample evaluation dataset, designed for testing semantic-aware privacy agents. 

    \noindent
    \textit{Evaluation and Analysis}: We evaluate our technique using real-world prompts and compare it against several baseline methods. Our analysis shows that prior work often prioritizes privacy at the expense of utility, while our approach reduces utility errors by a factor of 5 while still enhancing privacy.

\section{Background}

\textbf{Personally Identifiable Information} refers to any information that can be used to identify an individual, either directly or indirectly. PII is a key concept in privacy and data protection, as its exposure can lead to identity theft, fraud, and other security risks~\cite{healthcare8020133, 10.1145/1592665.1592668}. The definition of what constitutes PII can vary, but generally, it includes both direct and indirect identifiers~\cite{pilan2022text}. PII can be used in many malicious attacks, including the recent uptick in ``Pig Butchering" scams~\cite{Han16012025}. Attackers can use PII to lure unsuspecting victims into a false state of trust by using personal information against them, such as the identities of close family, workplace details, or contact information.

Prior studies have highlighted that identifying PII is a significant challenge~\cite{nadeau2009survey, pilan2022text}.
To address these challenges, most techniques rely on Named Entity Recognition (NER), a method used to identify and classify entities such as names, locations, and organizations within text. 
Existing methods, such as SpaCy~\cite{spacy2} and NLTK~\cite{bird2009natural}, leverage language models to perform NER. These models identify entities such as names, locations, organizations, and other relevant categories, classifying them into predefined labels like \colorbox[HTML]{fbd9d3}{\texttt{[PERSON]}} or \colorbox[HTML]{fbd9d3}{\texttt{[GPE]}} (Geopolitical Entity). Prior work has adopted spaCy as part of their pipeline to identify these named entities~\cite{chen2023hideseekhaslightweight}.






\section{Related Work}
Prior work has extensively studied the leakage of PII and privacy breaches through traffic originating from mobile devices and websites~\cite{starov2016you}. These studies highlight how data transmitted over networks can expose sensitive information. Such leaks enable third parties --- including advertisers, data brokers, and malicious actors --- to track users across different services. Such information can be correlated to build detailed user profiles without explicit consent.  Recent research has also highlighted similar privacy risks arising from LLMs~\cite{rocher2019estimating}. Today's growing reliance on LLMs introduces new challenges, as these models can inadvertently reveal sensitive information learned during training.
Many companies openly acknowledge using personal data to train their models~\cite{vox}, which increases potential privacy risks. For instance, there is research that demonstrates that models may expose sensitive details such as phone numbers, email addresses, full names, and location information~\cite{lee2023language}.
As a result, there has been recent work focused on mitigating these privacy concerns in language models~\cite{li2021large, chen2023hideseekhaslightweight}. These mitigation techniques often involve the use of differential privacy guarantees during the training pipeline. Additionally, efforts have been made to reduce PII leakage in language models specifically~\cite{zhao2022provably}. However, much of this work primarily focuses on privacy, often neglecting the preservation of utility and the semantic integrity of the generated outputs.
In contrast, our approach offers a balanced solution that effectively guards against unintentional PII leakage while still enabling users to interact with GPT-like models

\section{LOPSIDED Design}
\begin{figure}[t]
    \centering
    \includegraphics[width=3in]{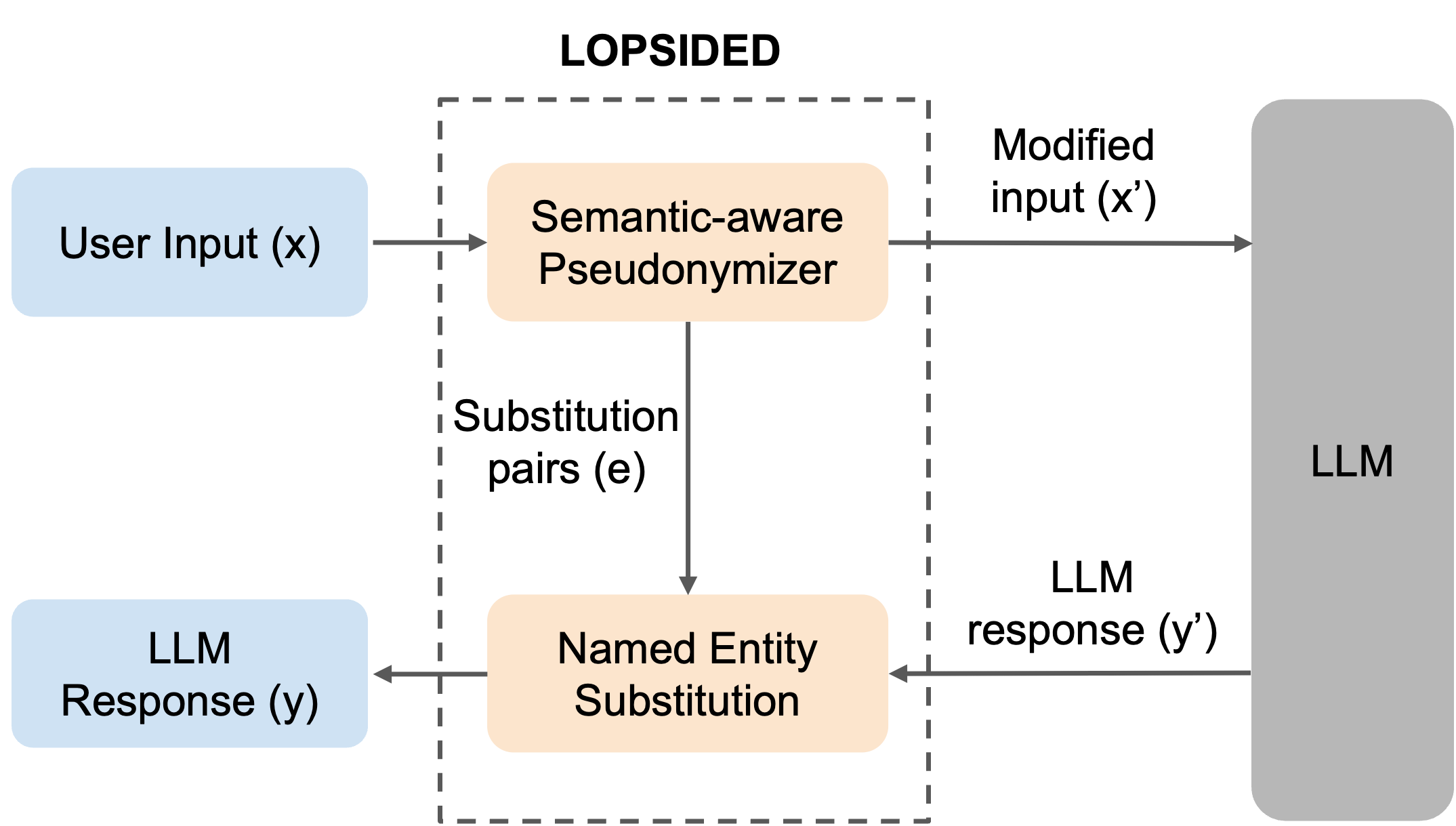}
    \caption{Overall workflow of LOPSIDED framework. }
    \label{fig:lopside}
\end{figure}
Figure~\ref{fig:lopside} illustrates the overall workflow of the LOPSIDED framework, which consists of two main components: \textit{semantic-aware pseudonymization} and the \textit{named entity substitution} module. Unlike prior methods, the semantic-aware pseudonymization module is designed to generate semantically appropriate replacement entities, referred to as \textit{pseudonyms}, for sensitive information while preserving the meaning of both the input prompt and the response derived from it. Specifically, when the user provides an input prompt, the semantic-aware pseudonymization module identifies and replaces private named entities, ensuring that this does not impact the overall response. The sensitive named entities are then stored locally for later use. The sanitized prompt is subsequently sent to the remote cloud provider, where a response is generated. Once returned, the named entity substitution module utilizes both the locally stored private named entity information and the generated response to produce the final output. As a result, the user receives a response that protects privacy while maintaining the utility of the original prompt.


\subsection{Semantic-aware Pseudonymization}

This component substitutes named entities within a user prompt with pseudonyms. 
Formally, let $x$ represent the original input prompt. We train a pseudonymization model $\mathcal{P}$ to generate an output consisting of a modified prompt $x'$ and a set of entity pairs $e=\{(e_{\text{orig}}, e_{\text{pseudo}})\}$, where $e_{\text{orig}}$ is a named entity identified in the original prompt $x$ and $e_{\text{pseudo}}$ is the corresponding pseudonym or replacement entity used in $x'$.

\subsubsection{Data collection for Pseudonymization}

\begin{figure}
\small
\begin{lstlisting}[language=json,firstnumber=1]
{
  "changed_entities": [
    {
      "explanation": "The name 'Raven' can be 
        any dog name...",
      "original_entity": "Raven",
      "new_entity": "Shadow"
    },
     ...
  ],
  "modified_prompt": "Write a short story about Shadow..."
}
\end{lstlisting}
    \caption{GPT-4o pseudonymization output.}
    \label{fig:pseudooutput}
\end{figure}

To train the model, we first collect a dataset of sentences containing sensitive named entities. Since manually modifying each sentence is both time-consuming and costly, we leverage GPT-4o to automate this process. We prompt GPT-4o to generate semantically appropriate replacement entities for the sensitive information in the sentences, resulting in a modified version of the input prompt. Figure~\ref{fig:pseudooutput} illustrates a sample response from GPT-4o. This approach allows us to create a supervised dataset, which we then use to distill the knowledge from GPT-4o into our model.


\subsubsection{Model Training.}
The pseudonymization model $\mathcal{P}$ is trained on the curated dataset $\mathcal{D}_{\text{pseudo}}$ to learn to generate contextually appropriate pseudonyms while preserving the semantic meaning of the prompt. Each training example consists of the original input $x$, the modified prompt $x'$ with pseudonymized entities, and the corresponding entity mapping $e$.

Formally, the model $\mathcal{P}$ is trained to maximize the likelihood of producing both the pseudonymized prompt and the entity mapping given the original input:
\begin{equation}
    \max_{\mathcal{P}} \mathbb{E}_{(x, e, x')\sim\mathcal{D}_{\text{pseudo}}} \log p_{\mathcal{P}}( e, x'| x)
 \end{equation}
where $e=\{(e_{\text{orig}}, e_{\text{pseudo}}\})$ is a set of named entity substitution pairs, and $x'$ is the modified prompt containing the pseudonymized entities. 

\subsection{Named Entity Substitution}
The key goal of the named entity substitution model $S$, also referred to as the replacement model for simplicity, is to reconstruct the original response $y$ as transparently as possible. Specifically, the model ensures that the final output is semantically faithful to the response that the remote LLM would have generated from the unmodified input $x$, while still protecting user privacy. 
To achieve this, note that after pseudonymization, the remote LLM produces a response $y'$ based on the modified input $x'$. Since $y'$ may contain pseudonyms rather than the original entities, the substitution model $S$ uses the stored reverse mapping $e'=\{(e_{\text{pseudo}}, e_{\text{orig}})\}$ to restore the original entities, ensuring that the output remains semantically consistent with what the LLM would have produced for $x$. 

For example, if the modified query \textit{``What is the weather in San Jose?"} yields $y'=$ \textit{``The weather in San Jose is sunny."}, the substitution model applies $e$ to produce $y=$ \textit{``The weather in Palo Alto is sunny."}. In this way, the user receives an accurate answer, while the sensitive entity never leaves the local environment.


Formally, let $x'$ represent the modified input and $y'$ denote the response generated by the remote LLM using $x'$. We train a substitution model $S$ to reconstruct the response $y$, which would have been generated by remote LLM from the original input $x$. The model $S$ takes the modified remote LLM response $y'$, and uses a set of named entity mapping $e'=\{(e_{\text{pseudo}}, e_\text{orig})\}$ to restore the original entities within a response $y$. 

\subsubsection{Data collection for substitution model}

We augment the dataset collected during the pseudonymization step by incorporating responses from GPT-4o. For each original input $x$ and its corresponding modified version $x'$, we query GPT-4o to generate both the original response $y$ (for $x$) and the modified response $y'$ (for $x'$). 

The process results in a dataset consisting of tuples of the form (original input, modified input, original response, modified response, named entity substitution pairs). We then use this dataset to train the substitution model $S$ to reconstruct the original response $y$ from the modified response $y'$ and named entity substitution pairs $e'$.








 
            




\subsubsection{Model training} We train the substitution model $S$ on the dataset $\mathcal{D}_\text{sub}$ to maximize the likelihood of reconstructing $y$ from $y'$ given $e'$. The objective is:
\begin{equation}
    \max_{\mathcal{S}}  \mathbb{E}_{(y, e', y')\sim\mathcal{D}_{\text{sub}}} \log p_{\mathcal{S}}( y| y', e')
 \end{equation}
where $e'=\{(e_{\text{pseudo}}, e_\text{orig})\}$ is a set of pairs that provides the pseudonym and its corresponding named entity. This setup ensures that $S$ learns to reintegrate entities in context, rather than performing naive string replacement.



\section{Dataset}
\begin{table}[t]
\small
\centering
\begin{tabular}{@{}lll@{}}
\toprule
Metric                        & Test Set & Training Set \\ \midrule
\# of Prompts                  & 866      & 2595         \\
\# of Entities                     & 1195     & 3696         \\
Entities per Prompt             & 1.38     & 1.42         \\
Avg \# of Word Tokens          & 49.38    & 49.03        \\
Avg Entity Length    & 6.80     & 7.24         \\
Max Ents in a Prompt          & 8        & 31           \\
\# Prompts Req. Review & 30       & N/A          \\
Rejections                    & 20       & N/A          \\ \bottomrule
\end{tabular}
\caption{Data statistics and validation summary.}
\label{tab:datastats}
\end{table}
\subsection{Data Collection}
We use the ShareGPT dataset, a publicly available dataset, consisting of 70K ChatGPT conversation history of  users~\cite{vicuna2023}. This dataset includes a wide variety of real user-generated conversations with AI systems, and is freely available.

We focus only on the first turn of conversation in the dataset, as expanding the context to include multiple turns would significantly increase the resources required for training the models. However, our approach is extendable to multi-turn conversations, and future work could explore how to efficiently handle longer context windows while maintaining the same level of privacy and semantic integrity.

As the majority of the samples in this corpus do not contain PII, we utilize Amazon Comprehend's PII Detection Service to filter for prompts that do. After running Comprehend on the dataset, the service flagged 3461 samples as containing PII. However, upon manually inspecting these samples, we found that the service is not always accurate and sometimes flags sentences that do not actually contain any PII. Nevertheless, we decided to retain these samples in the dataset. Later, when we use these prompts with GPT to identify named entities, the GPT responses for these flagged prompts contain no named entities, confirming that they do not actually contain PII. Table~\ref{tab:datastats} summarizes the key statistics of our dataset. 





We begin by tagging the named entities using spaCy, which also categorizes each entity (e.g., location, name). We then annotate whether the entities are relevant or irrelevant. We consider an entity relevant if substituting it would alter the meaning of the prompt or significantly impact the quality of the response. Irrelevant named entities can be safely replaced without affecting the response from an LLM. 

We developed a custom interface designed to streamline this process, which enables annotators to easily label entities. A local LLM runs in the background, allowing users to test how our privacy agent can impact the model's responses. On the one side of the interface, annotators can view the model’s original output without any privacy intervention. On the other side, they can observe the response generated  after replacing the entity with a randomly generated one of the same type. This comparison is provided as guidance to assist annotators in making informed decisions. We recruited three graduate students from our lab, who volunteered to assist with the annotation process. Each  was trained on how to use the interface and was instructed to label the entities in the dataset as relevant or irrelevant for PII. Given the substantial time and effort required for manual annotation, we limited the scope of the data annotation to the test dataset. This allowed us to evaluate how our approach performed in preserving privacy and semantic integrity. 

\subsection{Data Validation}
We validated our data using a majority voting approach. Specifically, if two out of the three annotators agreed on an entity being relevant, then it was classified as relevant; if two out of the three annotators agreed that it was irrelevant, it was considered irrelevant. Annotator agreement was around 75\%, with a free-marginal Kappa value of 0.64. Any number above 0 indicates better-than-chance agreement, while 1 indicates complete agreement. Our value shows that our annotators had substantial agreement~\cite{8d20e0b8-89d8-3d65-bcf5-8c19d56ec4ab}. 
\subsection{Data Analysis}

\begin{table}[t]
\small
\centering


\begin{tabular}{@{}lll@{}}
\toprule
\textbf{Type}         & \textbf{Relevant} & \textbf{Irrelevant} \\ \midrule
\textbf{Person}       & 228      & 363        \\
\textbf{Organization} & 160      & 54         \\
\textbf{Facility}     & 5        & 1          \\
\textbf{City/Country} & 267      & 23         \\
\textbf{Landmark}     & 26       & 4          \\
\textbf{Demographic}  & 62       & 2          \\ \bottomrule
\textbf{Total} & 748 & 447 
\end{tabular}

\caption{Statistics of our human annotated dataset. }
\label{tab:humandata}
\end{table}
Table~\ref{tab:humandata} highlights the key characteristics of the human-annotated dataset. We observe that, for each named entity category, the number of relevant and irrelevant samples varies. In general, relevant tags occur 1.6 times more frequently than irrelevant tags (Table~\ref{tab:humandata}), which aligns with the intuition that users typically include information that is relevant to the task. However, 37\% of the samples were deemed irrelevant, indicating that these named entities can be safely replaced without affecting LLM's response.


\section{Evaluation}

\subsection{Baseline Methods}

\textbf{Microsoft Presidio~\cite{MsPresidio}.} This data protection tool focuses on accurately detecting private information in text for anonymization or removal. It prioritizes privacy but does not consider the utility of the entities it removes. \textit{Presidio w/ Replacement} modifies the anonymizer by assigning identifiers to entities it replaces. This is stored as a mapping to the original, and is restored by the Presidio Deanonymizer. 

\textbf{Hide-and-Seek (HaS)~\cite{chen2023hideseekhaslightweight}.}
This privacy framework uses an LLM to anonymize and deanonymize prompts to LLM. The model focuses on privacy but does not consider semantic meaning. As a result, it may lead to responses that are semantically inconsistent to the user's intended query. \textit{HaS (fine-tuned)} is trained on our dataset to improve its performance in chat scenarios.



\subsection{Training}

We use a Gemma 2 2b model, and fine-tune it on our dataset. The model is fine-tuned using LoRa and 4 bit quantization on an A6000 GPU. 

\subsection{Metrics and Definitions}

\subsubsection{ROUGE and BLEU}

BLEU score measures how many words in the sample appear in the reference, which shows the precision. ROUGE score measures the degree of between the sample and reference, which shows recall. These methods are commonly used~\cite{graham2015re, blagec2022globalanalysismetricsused} in machine translation and summarization to evaluate the response quality.

\subsubsection{LLM-as-a-Judge}

 LLM-as-a-Judge~\cite{zheng2023judgingllmasajudgemtbenchchatbot} uses LLMs to rate a response on a scale, with guidelines on response quality, relativity, and fulfillment. We use GPT-4o-mini as our judge, due to its low cost. This allows us to achieve higher score accuracy than if using a smaller local model. As a reference, we have ran both Llama3-1b and GPT-4o-mini on our dataset without anonymization, and they were rated 8.17 and 9.32 respectively. We consider these the upper and lower bound for our experiments.

\subsubsection{Syntheticity}

We define a response to be synthetic if there are any perceived perturbation of said response on the user side of the pipeline. For example, any hallucinations caused by erroneously replaced entities would be evidence of a synthetic response. 

\subsubsection{Privacy and Utility Errors}
We evaluate the model using two complementary metrics: privacy and utility errors. \textbf{Privacy errors} occur when entities are \textit{not} replaced when they should have been. \textbf{Utility errors} are defined as the amount of relevant entities that were incorrectly replaced by the model.





\section{Results}

\subsection{Baseline Performance}
\begin{table*}[t]
\small
\centering
\begin{tabular}{@{}lcccccccc@{}}
\toprule
                          & ROUGE-1 & ROUGE-2 & ROUGE-L & BLEU-1 & BLEU-2 & BLEU-3 & BLEU-4  & LLM-as-a-Judge\\ \midrule 
LOPSIDED
&\textbf{ 0.796 }  &\textbf{ 0.625 }  & \textbf{0.654}   & \textbf{0.720}  & \textbf{0.641}  & \textbf{0.595}  & \textbf{0.564}  & \textbf{9.19} \\
HaS (finetuned)                      & 0.461   & 0.226   & 0.284   & 0.149  & 0.108  & 0.096  & 0.090  &  8.17\\
HaS                       & 0.149   & 0.102   & 0.125   & 0.139  & 0.129  & 0.124  & 0.121  & N/A\\
Presidio                  & 0.642   & 0.443   & 0.487   & 0.532  & 0.444  & 0.397  & 0.366  & N/A\\ 
Presidio w/ Repl & 0.655   & 0.454   & 0.497   & 0.541  & 0.453  & 0.405  & 0.374  & 7.29 \\ \bottomrule
\end{tabular}
\caption{Baseline performance comparisons. Bolded values are the highest scores. }
\label{tab:rougebleu}
\end{table*}

We begin by comparing our approach to baseline techniques. In our experiment, we modify the prompt and compare the output generated by the privacy agent to the output produced by GPT-4 alone, without any privacy interventions. To evaluate the performance of each approach, we use standard metrics such as ROUGE and BLEU scores, which assess the quality and similarity of the generated responses \cite{blagec2022globalanalysismetricsused}.

Table~\ref{tab:rougebleu} compares the performance of various privacy agents. As shown, LOPSIDED outperforms other techniques in terms of ROUGE/BLEU scores. In general, models that were not fine-tuned exhibit lower performance. Notably, LOPSIDED achieves higher scores, indicating that its responses are closer to the ground truth (i.e., the original, unmodified response) compared to other baseline techniques. Additionally, models that have been fine-tuned tend to have lower scores, further highlighting the effectiveness of LOPSIDED in preserving semantic integrity of the LLM response. Also included in this table are the results from our LLM-as-a-Judge evaluation. This evaluation gives a a reasonably accurate estimate to what a human evaluation of our framework's output would be. LOPSIDED not only has the highest score of all the frameworks, but it also scores only .13 points lower than our upper bound. 






\subsection{Privacy and Utility Evaluation}
Next, we evaluate the performance of our substitution model in balancing privacy and utility. For this evaluation, we use the human-annotated test set, which contains labels indicating whether each entity in a prompt is relevant or irrelevant. By comparing the output of the substitution model with these labels, we can measure how well it maintains the utility of the response by ensuring that relevant entities remain intact, while effectively substituting irrelevant or private entities. 




\begin{figure}
    \centering
    \includegraphics[width=\linewidth]{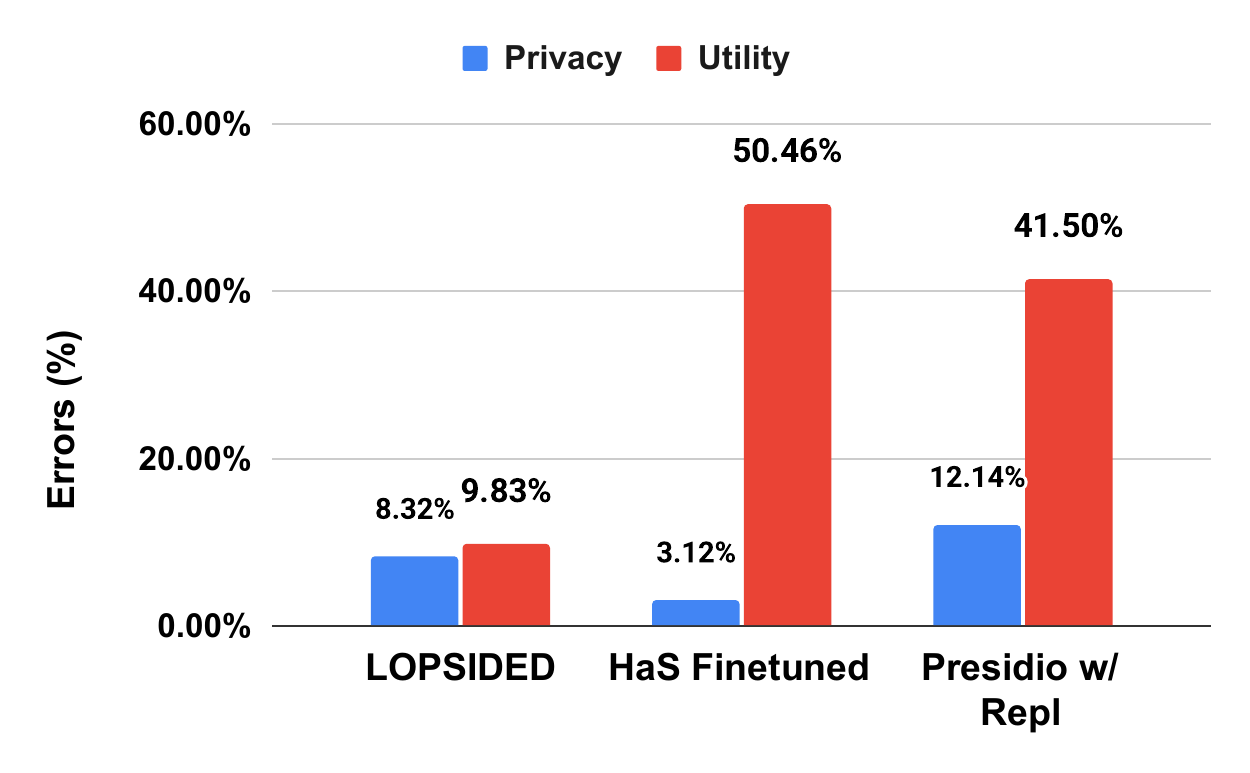}
    
    \caption{Privacy and utility error rate comparisons.}
    \label{fig:humaneval}
\end{figure}

Figure~\ref{fig:humaneval} shows that HaS achieves a low privacy error of 3\% because it primarily focuses on substituting all named entities, regardless of their relevance. However, this comes at the cost of higher utility errors. LOPSIDED has only a slightly higher privacy error of 8\%, but it achieves 5$\times$ fewer utility-related errors, demonstrating its ability to preserve relevant entities while still protecting private information. Compared to Presidio, LOPSIDED achieves lower privacy \textit{and} utility errors.

\subsection{Text Syntheticity Detection}
\begin{table}[t]
\centering
\small
\begin{tabular}{@{}lcc@{}}
\toprule
Framework               & Detectability& Detectability  \\
               & (Avg) & (Final) \\\midrule
LOPSIDED                & \textbf{46.59\%}                     & \textbf{44.88\%}                   \\
HaS Finetuned           & 71.84\%                     & 83.84\%                   \\
HaS                     & 85.65\%                     & 87.73\%                   \\
Presidio                & 60.43\%                     & 62.19\%                   \\
Presidio w/ Repl & 51.36\%                     & 48.63\%                   \\ \bottomrule
\end{tabular}
\caption{Syntheticity detection scores.}
\label{tab:syntheticity}
\end{table}
Similar to Yermilov et al, we conduct a text syntheticity detection experiment. This analysis is necessary because pseudonymization can disrupt the relationships between named entities and their surrounding context, potentially leading to inconsistencies in downstream tasks. We combine responses from both pseudonymized and original texts and train a classification model using \texttt{bert-base-uncased} to distinguish between the two. A high classification accuracy indicates that pseudonymization introduces detectable artifacts, whereas a low accuracy suggests that modified texts closely resemble their original counterparts. Table~\ref{tab:syntheticity} presents the text syntheticity classification scores for different techniques. As shown, LOPSIDED achieves the lowest classification score, indicating that its pseudonymized texts are the most similar to the original ones. 

\section{Conclusion}

LOPSIDED introduces a novel framework for pseudonymizing LLM API prompts while preserving the utility of the user's request. The methods we use introduce the key contribution of selectively changing private entities, alleviating the destructive effects of all-or-nothing solutions, while still enhancing the user's privacy protections. When comparing other state of the art privacy preserving methods, LOPSIDED protects the utility of named entities at a significantly greater rate. At the same time, we replace the irrelevant private entities at a similar, or better, rate. These privacy protections help users remain safer in a rapidly scaling world of AI systems. To support our claims, we present an evaluation dataset validated by human annotators to assess the effectiveness and utility of privacy agents. This dataset serves as a benchmark for evaluating similar privacy techniques and demonstrates the strengths of our approach. Our framework is extendable, with the ability for future researchers to extend the platform to ensure robustness in multi-turn scenarios, and more intricate replacement strategies. Supporting datasets and models can be found on our GitHub repository: \textit{https://github.com/jmseren/LOPSIDED}.

{\bf Acknowledgments.} This research was supported in part by the University of Pittsburgh Center for Research Computing and Data, RRID:SCR\_022735, through the resources provided. Specifically, this work used the H2P cluster, which is supported by NSF award number OAC-2117681.

\bibliographystyle{IEEEtran}
\bibliography{latex/custom}

\end{document}